\documentclass[11pt,a4paper]{article}
\usepackage[table,dvipsnames]{xcolor}
\usepackage[hyperref]{emnlp2019}
\usepackage{times}
\usepackage{latexsym}

\usepackage{url}
\usepackage{graphicx}               
\usepackage{tabularx}               
\newcolumntype{C}{>{\centering\arraybackslash}X}
\usepackage{multirow}               
\usepackage{diagbox}                
\usepackage{hhline}                 
\usepackage{color}                  
\usepackage{amsmath}                
\usepackage{amssymb}                
\usepackage{mathtools}              

\usepackage{subfigure}

\newcommand{\vpara}[1]{\vspace{0.01in}\noindent\textbf{#1 }}

\definecolor{RoseQuartzBg}{HTML}{F7CAC9}
\definecolor{RoseQuartz}{HTML}{F5A798}
\definecolor{Serenity}{HTML}{92A8D1}

\usepackage{enumitem}

\usepackage{arydshln}               
\definecolor{themered}{HTML}{FF8375}

\usepackage{xparse}
\NewDocumentCommand{\heng}{ mO{} }{\textcolor{OrangeRed}{\textsuperscript{\textit{Heng}}\textsf{\textbf{\small[#1]}}}}
\NewDocumentCommand{\yixin}{ mO{} }{\textcolor{blue}{\textsuperscript{\textit{Yixin}}\textsf{\textbf{\small[#1]}}}}

\NewDocumentCommand{\method}{ mO{} }{#1}

\aclfinalcopy 

\title{Low-Resource Name Tagging Learned with Weakly Labeled Data}

\author{Yixin Cao$^1$ \quad Zikun Hu$^1$ \quad Tat-Seng Chua$^1$\\
\textbf{Zhiyuan Liu$^2$ \quad Heng Ji$^3$}\\
$^1$School of Computing, National University of Singapore, Singapore\\
$^2$Department of CST, Tsinghua University, Beijing, China\\
$^3$Department of CS, University of Illinois Urbana-Champaign, U.S.A.\\
{\tt caoyixin2011@gmail.com,zikunhu@u.nus.edu,dcscts@nus.edu.sg} \\
{\tt liuzy@tsinghua.edu.cn,hengji@illinois.edu} \\
}

\date{}
\begin{document}
\maketitle
\begin{abstract}
  Name tagging in low-resource languages or domains suffers from inadequate training data. Existing work heavily relies on additional information, while leaving those noisy annotations unexplored that extensively exist on the web. In this paper, we propose a novel neural model for name tagging solely based on weakly labeled (WL) data, so that it can be applied in any low-resource settings. To take the best advantage of all WL sentences, we split them into high-quality and noisy portions for two modules, respectively: (1) a classification module focusing on the large portion of noisy data can efficiently and robustly pretrain the tag classifier by capturing textual context semantics; and (2) a costly sequence labeling module focusing on high-quality data utilizes Partial-CRFs with non-entity sampling to achieve global optimum. Two modules are combined via shared parameters. Extensive experiments involving five low-resource languages and fine-grained food domain demonstrate our superior performance (6\% and 7.8\% F1 gains on average) as well as efficiency\footnote{Our project can be found in \url{https://github.com/zig-kwin-hu/Low-Resource-Name-Tagging}.}.
\end{abstract}

\section{Introduction}
Name tagging\footnote{Someone may call it Named Entity Recognition (NER).} is the task of identifying the boundaries of entity mentions in texts and classifying them into the pre-defined entity types (e.g., person). It serves as a fundamental role as providing the essential inputs for many IE tasks, such as Entity Linking~\citep{cao2018neural} and Relation Extraction~\cite{lin2017neural}.

\begin{figure}[htb]
  \centerline{\includegraphics[width=0.47\textwidth]{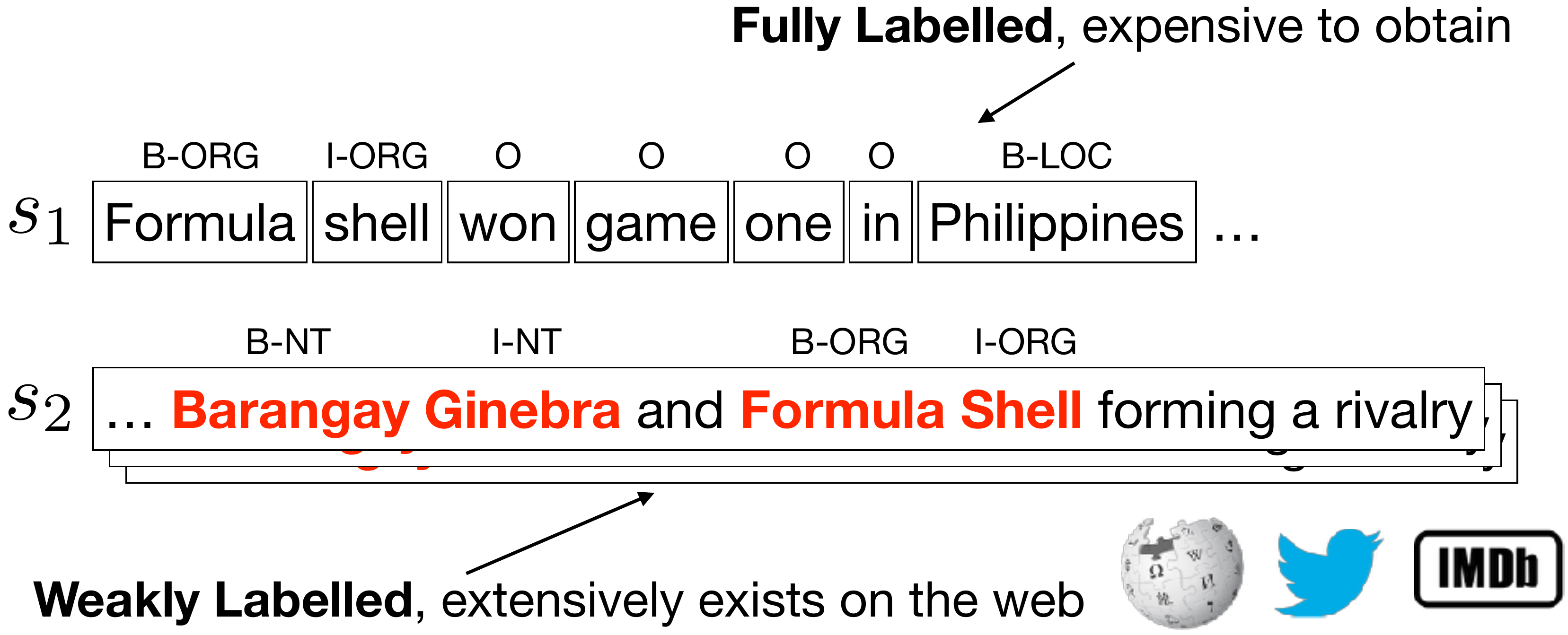}}
  \caption{Example of weakly labeled data. B-NT and I-NT denote incomplete labels without types.}
  \label{fig:example}
\end{figure}

Many recent methods utilize a neural network (NN) with Conditional Random Fields (CRFs)~\cite{lafferty2001conditional} by treating name tagging as a sequence labeling problem~\cite{lample2016neural}, which has became a basic architecture due to its superior performance. Nevertheless, NN-CRFs require exhaustive human efforts for training annotations, and may not perform well in low-resource settings~\cite{ni2017weakly}. Many approaches thus focus on transferring cross-domain, cross-task and cross-lingual knowledge into name tagging~\cite{yang2016transfer,peng2016improving,mayhew2017cheap,pan2017cross,lin2018multi,xie2018neural}. However, they are usually limited by the extra knowledge resources that are effective only in specific languages or domains.

Actually, in many low-resource settings, there are extensive noisy annotations that naturally exist on the web yet to be explored~\cite{ni2017weakly}. In this paper, we propose a novel model for name tagging that maximizes the potential of weakly labeled (WL) data. As shown in Figure~\ref{fig:example}, $s_2$ is weakly labeled, since only \textit{Formula shell} and \textit{Barangay Ginebra} are annotated, leaving the remaining words unannotated.

WL data is more practical to obtain, since it is difficult for people to accurately annotate those entities that they do not know or are not interested in. We can construct them from online resources, such as the anchors in Wikipedia. However, the following natures of WL data make learning name tagging from them more challenging:


\textbf{Partially-Labeled Sequence} Automatically derived WL data does not contain complete annotations, thus can not be directly used for training. \citet{ni2017weakly} select the sentences with highest confidence, and assume missing labels as O (i.e., non-entity), but it will introduce a bias to recognize mentions as non-entity. Another line of work is to replace CRFs with Partial-CRFs~\cite{tackstrom2013token}, which assign unlabeled words with all possible labels and maximize the total probability~\cite{yang2018distantly,shang2018learning}. However, they still rely on seed annotations or domain dictionaries for high-quality training.

\textbf{Massive Noisy Data} WL corpora are usually generated with massive noisy data including missing labels, incorrect boundaries and types. Previous work filtered out WL sentences by statistical methods~\cite{ni2017weakly} or the output of a trainable classifier~\cite{yang2018distantly}. However, abandoning training data may exacerbate the issue of inadequate annotation. Therefore, maximizing the potential of massive noisy data as well as high-quality part, yet being efficient, is challenging.

To address these issues, we first differentiate noisy data from high-quality WL sentences via a lightweight scoring strategy, which accounts for the annotation confidence as well as the coverage of all mentions in one sentence. To take best advantages of all WL data, we then propose a unified neural framework that solves name tagging from two perspectives: sequence labeling and classification for two types of data, respectively.

Specifically, the classification module focuses on noisy data to efficiently pretrain the tag classifier by capturing textual context semantics. It is trained only using annotated words without noisy unannotated words, and thus it is robust and efficient during training. The costly sequence labeling module is to achieve sequential optimum among word tags. It further alleviates the burden of seed annotations in Partial-CRFs and increases randomness via Non-entity Sampling strategy, which samples O words according to some linguistic natures. These two modules are combined via shared parameters. Our main contributions are as follows:

  \begin{figure*}[htb]
    \centerline{\includegraphics[width=0.95\textwidth]{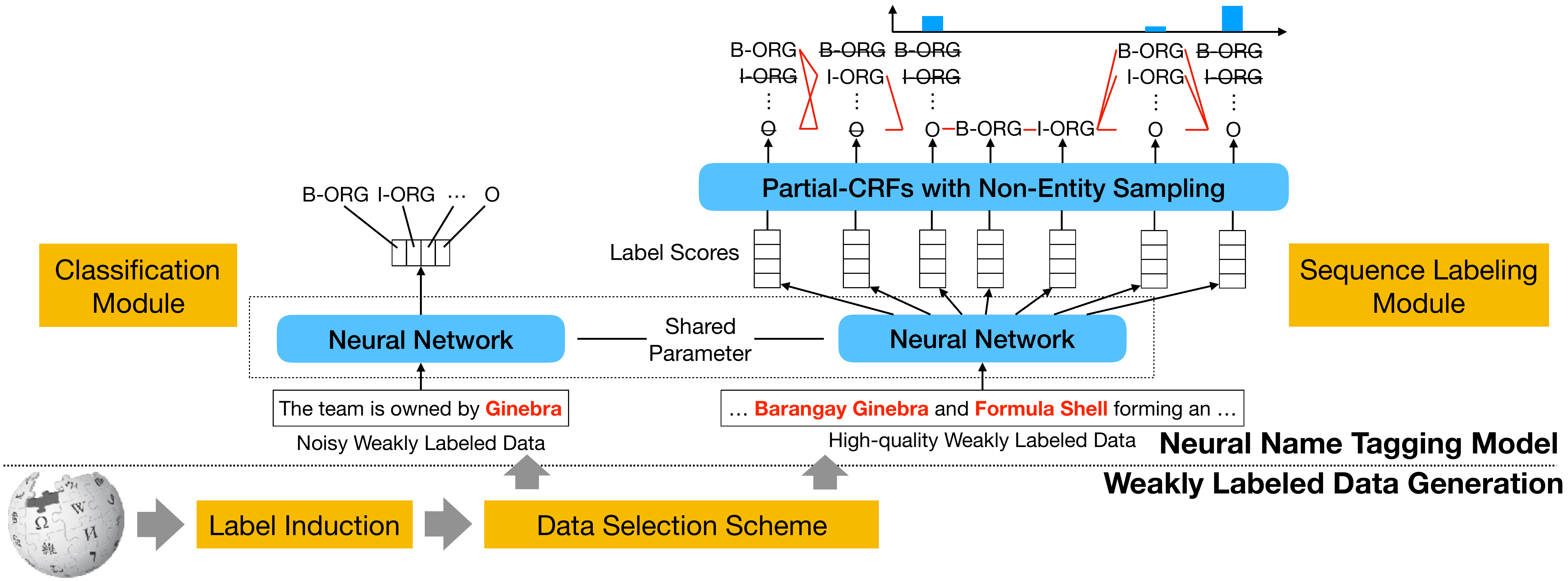}}
    \caption{Framework. Rectangles denote the main components for two steps, and rounded rectangles consist of two modules of the neural model. In input sentences, bold fonts denote labeled words, and at the top is corresponding outputs. We use Partial-CRFs to model all possible label sequences (red paths from left to right by picking up one label per column) controlled by non-entity sampling (strikethrough labels according to the distribution). We replace ``UN'' and ``x-NT'' label with corresponding possible labels to clarify the principle of PCRF.}
    \label{fig:framework}
  \end{figure*}

\begin{itemize}
  \item We propose a novel neural name tagging model that merely relies on WL data without feature engineering. It can thus be adapted for both low-resource languages and domains, while no previous work deals with them at the same time.
  \item We consider name tagging from two perspectives of sequence labeling and classification, to efficiently take the best advantage of both high-quality and noisy WL data.
  \item We conduct extensive experiments in five low-resource languages and a fine-grained domain. Since few work has been done in two types of low-resource settings simultaneously, we arrive at two types of baselines from state-of-the-art methods. Our model achieves significant improvements (6\% and 7.8\% F1 on average), yet being efficient demonstrated in further ablation studies.
\end{itemize}

\section{Related Work}
\label{sec:rw}
Name tagging is an fundamental task of extracting entity information, which shall benefit many applications, such as information extraction~\cite{zhang2017xlink,kuang2019improving,cao2019multi} and recommendation~\cite{wang2019explainable,cao2019unifying}. It can be treated as either a multi-class classification problem~\cite{hammerton2003named,xu2017local} or a sequence labeling problem~\cite{collobert2011natural}, but very little work combined them together. The difference between them mainly lies in whether the method models sequential label constraints, which have been demonstrated effective in many NN-CRFs models~\cite{lample2016neural,ma2016end,chiu2016named}. However, they require a large amount of human annotated corpora, which are usually expensive to obtain.

The above issue motivates a lot of work on name tagging in low-resource languages or domains. A typical line of effort focuses on introducing external knowledge via transfer learning~\cite{fritzler2018few,hofer2018few}, such as the use of cross-domain~\cite{yang2016transfer}, cross-task~\cite{peng2016improving,lin2018multi} and cross-lingual resources~\cite{ni2017weakly,xie2018neural,zafarian2015semi,zhang2016name,mayhew2017cheap,tsai2016cross,feng2018improving,pan2017cross}. Although they achieve promising results, there are a large amount of weak annotations on the Web, which have not been well studied~\cite{nothman2008transforming,ehrmann2011building}. \citet{yang2018distantly,shang2018learning} utilized Partial-CRFs~\cite{tackstrom2013token} to model incomplete annotations for specific domains, but they still rely on seed annotations or a domain dictionary. Therefore, we aim at filling the gap in low-resource name tagging research by using only WL data, and adapt it to arbitrary low-resource languages or domains, which can be further improved by the above transfer-based methods.

\section{Preliminaries and Framework}
\subsection{Preliminaries}
We formally define the name tagging task as follows: given a sequence of words $X=\langle x_1, \cdots, x_i,\cdots, x_{|X|} \rangle$, it aims to infer a sequence of labels $Y=\langle y_1,\cdots,y_i,\cdots,y_{|X|}\rangle$, where $|X|$ is the length of the sequence, $y_i\in \mathcal{Y}$ is the label of the word $x_i$, each label consists of the boundary and type information, such as B-ORG indicating that the word is \textbf{B}egin of an \textbf{ORG}anization entity.
To make notations consistent, we use $\tilde{\mathcal{Y}}=\mathcal{Y}\bigcup \{\text{UN,B-NT,I-NT}\}$ to denote the label set of WL data, where UN indicates that the word is unlabeled, and NT denote only the type is unlabeled. In other words, the word with UN may be any one of the label in $\mathcal{Y}$, and the word with NT may be any type. We define $\tilde{\mathcal{Y}}$ for notation clarity.

To deal with the issue of limited annotations, we construct WL data $\mathcal{D}=\{(X,\tilde{Y})\}$ based on Wikipedia anchors and taxonomy, where $\tilde{Y}=\langle \tilde{y}_1,\cdots,\tilde{y}_i,\cdots,\tilde{y}_{|X|}\rangle$ and $\tilde{y}_i\in\tilde{\mathcal{Y}}$. An \textbf{anchor} $\langle m,e\rangle\in \mathcal{A}$ links a mention $m$ to an entity $e\in E$, where $m$ contains one or several consecutive words of length $|m|$. Particularly, we define $\mathcal{A}(X)$ as the set of anchors in $X$. Most entities are mapped to hierarchically organized categories, namely \textbf{taxonomy} $\mathcal{T}$, which provides category information $C=\{c\}$. We define $C(e)$ as the category set of $e$, and $\mathcal{T}_{\downarrow}(c)$ as the children of $c$.

\subsection{Framework}
The goal of our method is to extract WL data from Wikipedia and use them as training corpora for name tagging. As shown in Figure~\ref{fig:framework}, there are two steps in our framework:

\textbf{Weakly Labeled Data Generation} generates as many WL data as possible for higher tagging recall. It contains two components of label induction and data selection scheme. First, the label induction assigns each word a label based on Wikipedia anchors and taxonomy. Then, the data selection scheme computes the quality scores for the WL sentences by considering the coverage of mentions as well as the label confidence. According to the scores, we split the entire set into two parts: a small set of high-quality data for the sequence labeling module, and a large amount of noisy data for the classification module.

\textbf{Neural Name Tagging Model} aims at efficiently and robustly utilizing both high-quality and noisy WL data, ensuring satisfying tagging precision. It is to make best use of labeled words via the sequence labeling module and the classification module. More specifically, we pre-train the classification module to capture the textual context semantics from massive noisy data, and then the sequence labeling module further fine-tunes the shared neural networks using a Partial-CRFs layer with Non-Entity Sampling.

\section{Weakly Labeled Data Generation}
\label{sec:wlg}
Existing methods use Wikipedia~\cite{ni2017weakly,pan2017cross,geiss2017neckar} to train an extra classifier to predict entity categories for name tagging training. Instead, we aim at lowering the requirements of additional resources in order to support more low-resource settings. We thus utilize a lightweight strategy to generate WL data including label induction and data selection scheme.

\subsection{Label Induction}
\label{sec:li}
Given a sentence $X$ including anchors $\mathcal{A}(X)$ and taxonomy $\mathcal{T}$, we aim at inducing a label $\tilde{y}\in\tilde{\mathcal{Y}}$ for each word $x\in X$. Obviously, the words outside of anchors should be labeled with UN, indicating that it is unlabeled and could be O or unannotated mentions. For the words in an anchor $\langle m,e\rangle$, we label it according to the entity categories. 
For example, words \textit{Formula} and \textit{Shell} (Figure~\ref{fig:example}) in $s_2$ are labeled as B-ORG and I-ORG, respectively, because mention \textit{Formula Shell} is linked to entity \textit{Shell Turbo Chargers}, which belongs to category \textit{Basketball teams}. We trace it along the taxonomy $\mathcal{T}$: \textit{Basketball teams}$\rightarrow$\textit{...}$\rightarrow$\textit{Organizations}, and find that it is a child of \textit{Organizations}. According to a manually defined mapping $\Gamma(\mathcal{Y})\rightarrow C$ (e.g., $\Gamma(ORG)=$\textit{Organizations}), we denote all the classes and their children with the same type (e.g., ORG).

However, there are two minor issues. First, for the entities without category information $C(e)=\emptyset$, we label them as B-NT or I-NT, indicating that they have no type information. Second, for the entities referring to multiple categories, we induce labels that maximizes the conditional probability:

\begin{equation}
  \label{equ:etype}
  \underset{y^*}{\text{argmax}}\ p(y^*|C(e))=\frac{\sum_{c\in C(e)}\mathbf{1}(c\in \mathcal{T}_{\downarrow}(\Gamma(y^*)))}{|C(e)|}
\end{equation}
where $\mathbf{1}(\cdot)$ indicates $1$ if holds true, otherwise 0.

By doing so, we obtain a set of WL sentences $\mathcal{D}=\{(X,\tilde{Y})\}$. However, the induction process may introduce incorrect boundaries and types of labels due to the crowdsourcing nature of source data. We thus design a data selection scheme to deal with the issues.

\subsection{Data Selection Scheme}
\label{sec:dss}
Following~\citet{ni2017weakly}, we compute quality scores for sentences to distinguish high-quality and noisy data from two aspects: the annotation confidence and the annotation coverage.

The annotation confidence measures the likelihood of the text spans being mentions (i.e., correctness of boundaries), and being assigned with the types. We define it as follows:

\begin{equation}
  q(X,\tilde{Y})=\frac{\underset{(x_i,\tilde{y}_i)}{\sum} \mathbf{1}(\tilde{y}_i\in \mathcal{Y}) p(\tilde{y}_i|C(e))p(C(e)|x_i)}{|X|}
\end{equation}
where $p(C(e)|x_i)$ is the conditional probability of $x_i$ linking to an entity belong to category $C(e)$, we compute it based on its statistical frequency among Wikipedia anchors.

The annotation coverage measures to which ratio the words are being labeled in the sentence:

\begin{equation}
  n(X,\tilde{Y})=\frac{\underset{(x_i,\tilde{y}_i)}{\sum} \mathbf{1}(\tilde{y}_i\in \mathcal{Y})}{|X|}
\end{equation}
We select high-quality sentences $\mathcal{D}^{hq}$ satisfying:

\begin{equation}
  q(X,\tilde{Y})\geq \theta_q; n(X,\tilde{Y})\geq \theta_n
\end{equation}
where $\theta_q$ and $\theta_n$ are the hyperparameters. Thus, the remaining sentences are noisy $\mathcal{D}^{noise}$.

For example (Figure~\ref{fig:framework}), the sentence \textit{... Barangay Ginebra and Formula Shell ...} is high-quality, and \textit{The team is owned by Ginebra} is noisy. This is because there are more anchors that link \textit{Formula Shell} to an organization entity and the anchors within the sentence account for a large proportion, leading to a higher quality score. Note that \textit{Barangay} and \textit{Ginebra} are labeled with B-NT and I-NT, indicating the type information is missing. Our model may learn the textual semantics for classifying \textit{Ginebra} to ORG from the noisy sentence, where \textit{Ginebra} is labeled with B-ORG.

\section{Neural Name Tagging Model}
Our neural model contains two modules that share the same NN architecture except the Partial-CRFs layer.
Given $\mathcal{D}^{hq}$ and $\mathcal{D}^{noise}$, we first pre-train the classification module using massive noisy data $\mathcal{D}^{noise}$ to efficiently capture the textual semantics. Then, we use the sequence labeling module to fine-tune the classification module on high-quality data $\mathcal{D}^{hq}$ by considering the transitional constraints among sequential labels.

\subsection{Sequence Labeling Module}
\label{sec:slm}
Before describing the NN of the classification module, we first introduce the sequence labeling module. Different from conventional NN-CRFs models, we utilize the Partial CRFs layer to maximize the probability of all possible sequential labels for the sentence with transitional constraints, where the probability of missing word labels is controlled by non-entity sampling.

\subsubsection*{Partial-CRFs}
\label{sec:pCRFs}
Partial-CRFs (PCRFs) was first proposed in the field of Part-of-Speech Tagging~\cite{tackstrom2013token}. It can be trained when the coupled word and label constraints provide only a partial signal by assuming that the uncoupled words may refer to multiple labels. Given $(X,\tilde{Y})$, we traverse all possible labels $\mathcal{Y}$ for each unannotated word $\{x_i|\tilde{y}_i\in\text{UN,B-NT,I-NT}\}$ (e.g., the red paths in Figure~\ref{fig:framework}), and compute the total probability of possible fully labeled sentences $\mathcal{Y}(X,\tilde{Y})=\{(X,Y)\}$:

\begin{equation}
  p(\tilde{Y}|X)=\sum^{\mathcal{Y}(X,\tilde{Y})}_{(X,Y)} p(Y|X)
\end{equation}
where $p(Y|X)=\text{softmax}(s(X,Y))$, the same as in CRFs, and the score function $s(X,Y)$ is:

\begin{equation}
  \label{equ:CRFs}
  s(X,Y)=\sum_{i=0}^{|X|} A_{y_i,y_{i+1}}+\sum_{i=1}^{|X|} P_{x_i,y_i}
\end{equation}
where $P_{x_i,y_i}$ is the score indicating how possible $x_i$ is labeled with $y_i$, which is defined as the output of NN and will be detailed in the next section. 
$A_{y_i,y_{i+1}}$ is the transition score from label $y_i$ to $y_{i+1}$ that is learned in this layer.

Instead of the single correct label sequence in CRFs, the loss function of partial-CRFs is to minimize the negative log-probability of ground truth over all possible labeled sequences:

\begin{equation}
    \mathcal{L} = -\sum^{\mathcal{D}^{nq}}_{(X,\tilde{Y})}\log p(\tilde{Y}|X)
\end{equation}

\subsubsection*{Non-entity Sampling}
\label{sec:nes}
A crucial drawback of using partial CRFs for WL sentences is that there are no words labeled with O (i.e., non-entity words) for training (Section~\ref{sec:nsr}).

To further alleviate the reliance on seed annotations, we introduce non-entity sampling that samples O from unlabeled words as follows:

\begin{equation}
  \label{equ:nes}
  \begin{aligned}
    p(y_i&=\text{O}|x_i,\tilde{y}_i=\text{N}) = \\
    &\frac{\alpha}{3}(\lambda_1 f_1 + \lambda_2 (1-f_2) + \lambda_3 f_3)
  \end{aligned}
\end{equation}
where $\alpha$ is non-entity ratio to balance how many unlabeled words are sampled as O, we set $\alpha=0.9$ in experiments according to~\citet{augenstein2017generalisation}. Weighting parameters satisfy $0\leq\lambda_1,\lambda_2,\lambda_3\leq 1$, and $f_1,f_2,f_3$ are feature scores. We define $f_1=\mathbf{1}(x_i\text{ adjoins an entity})$, which implies that the words around a mention are possible to be O; $f_2$ is the ratio of the number of $x_i$ labeled with entities to its total occurrences, reflecting how frequent a word is in a mention; and $f_3=tf*df$, where $tf$ is term frequency and $df$ is document frequency in Wikipedia articles.

As shown in Figure~\ref{fig:framework}, three words \textit{and}, \textit{forming} and \textit{an} are labeled with N since they are outside of anchors. During training, they should be regarded as all labels of $\mathcal{Y}$ in Partial CRFs, while we sample some of them as O words according to Equation~\ref{equ:nes}. Thus, \textit{and} and \textit{an} are instead treated as O words, because they do not appear in any anchor, and are too general due to a high $f_3$ value.

\subsection{Classification Module}
\label{sec:cm}
To efficiently utilize the noisy WL sentences, this module regards name tagging as a multi-label classification problem. On one hand, it predicts each word's label separately, naturally addressing the issue of inconsecutive labels. On the other hand, we only focus on the labeled words, so that the module is robust to the noise since most noise arises from the unlabeled words, and enjoy an efficient training procedure.

Formally, given a noisy sentence $(X,\tilde{Y})\in\mathcal{D}^{noise}$, we classify words $\{x_i|\tilde{y}_i\in\mathcal{Y}\}$ by capturing textual semantics within the context. Independently of languages and domains, we combine the character and word embeddings for each word, then feed them into an encoder layer to capture contextual information for the classification layer.


\subsubsection*{Character and Word Embeddings}
As inputs, we introduce character information to enhance word representations to improve the robustness to morphological and misspelling noise following~\cite{ma2016end}. Concretely, we represent a word $\mathbf{x}$ by concatenating word embedding $\mathbf{w}$ and Convolutional Neural Networks (CNN)~\cite{lecun1989backpropagation} based character embedding $\mathbf{c}$, which is obtained through convolution operations over characters in a word followed by max pooling and drop out techniques.

\subsubsection*{Encoder Layer}

Given an arbitrary length of sentence $X$, this component encodes the semantics of words as well as their compositionality into a low-dimensional vector space. The most common encoders are CNN, Long-Short Term Memory (LSTM)~\cite{hochreiter1997long} and Transformer~\cite{vaswani2017attention}. We use the bi-directional LSTM (Bi-LSTM) due to its superior performance. We discuss it in Section~\ref{sec:lrl}.

Bi-LSTM~\cite{graves2013speech} has been widely used in modeling sequential words, so as to capture both past and future input features for a given word. It stacks a forward LSTM and a backward LSTM, so that the output of a word $x_i$ is $\mathbf{h}_i=[\overleftarrow{\mathbf{h}}_i;\overrightarrow{\mathbf{h}}_i]$, where $\overrightarrow{\mathbf{h}}_i=\text{LSTM}(X_{1:i})$ and $\overleftarrow{\mathbf{h}}_i=\text{LSTM}(X_{i:|X|})$.

\subsubsection*{Classification Layer}
The classification layer makes independent labeling decisions for each word, so that we can only focus on labeled words, while robustly and efficiently skip the noisy unlabeled words.

In this layer, we estimate the score $P_{x_i,y_i}$ (Equation~\ref{equ:CRFs}) for word $x_i$ being the label $y_i$. We use a fully connected layer followed by softmax to output a probability-like score:

\begin{equation}
  \label{equ:class}
  P_{x_i,y_i}=Softmax(W\mathbf{h}_i+b)
\end{equation}
where $W\in\mathbb{R}^{|\mathcal{Y}|}$. Note that we have no training instance for O words. Thus, we also use the non-entity sampling (Section~\ref{sec:nes}). Given $(X,\tilde{Y})\in\mathcal{D}^{noise}$, this module is trained to minimize cross-entropy of the predicted and ground truth:

\begin{equation}
  \mathcal{L}_c = -\sum^{\mathcal{D}^{noise}}_{(X,\tilde{Y})} \mathbf{1}(\tilde{y}_i\in\mathcal{Y}) \tilde{y}_i \log P_{x_i,y_i}
\end{equation}

\subsection{Training and Inference}
To distill the knowledge derived from noisy data, we first pre-train the classification module, then share the overall NN with the sequence labeling module. If we choose a loose threshold $\theta_p$ and $\theta_n$, there is no noisy data and our model shall degrade to the sequential model without the pre-trained classifier. When the threshold is strict, there is no high-quality data and our model will degrade to the classification module only (Section~\ref{sec:ea}).

For inference, we use the sequence labeling module to predict the output label sequence with the largest score as in Equation~\ref{equ:CRFs}.

\section{Experiment}
We verify our model using five low-resource languages and a specific domain. Furthermore, we investigate the impacts of the main components as well as hyperparameters in the ablation study.

\subsection{Experiment Settings}

\vpara{Datasets}
Since most datasets on low-resource languages are not publicly available, we use Wikipedia data as the ``ground truth'' following~\citet{pan2017cross}. Thus, we can test name tagging in low-resource languages as well as domains. We choose five languages: Welsh, Bengali, Yoruba, Mongolian and Egyptian Arabic (or CY, BN, YO, MN and ARZ for short), at different low-resource levels, and select 3 types: Person, Location and Organization. For food domain, we reorganized the entities in Wikipedia category “Food and drink” into 5 types: Drinks, Meat, Vegetables, Condiments and Breads, for name tagging, and extract sentences containing those entities from all English Wikipedia articles for as many data as possible.

\begin{table}[htp]
  \small
  \begin{center}
    \begin{tabular}{|c|cc|cc|} \hline 
      & \multicolumn{2}{c|}{\textbf{Train}} & \multicolumn{2}{c|}{\textbf{Test}} \\
        & \textbf{\#Sent} & \textbf{\#Ment} & \textbf{\#Sent} & \textbf{\# Ment} \\ \hline 
      CY & 106,541 & 146,524 & 1,193 & 3,256 \\ 
      BN & 66,915 & 127,932 & 870 & 2838 \\ 
      YO & 36,548 & 10,405 & 77 & 232 \\
      MN & 19,250 & 27,820 & 173 & 439 \\ 
      ARZ & 18,700 & 28,928 & 195 & 377 \\ \hline\hline
      Food Domain & 27,798 & 32,155 & 207 & 253 \\ \hline
      Drinks & 8,615 & 9,218 & 62 & 67 \\
      Meat & 7,685 & 8,841 & 53 & 68 \\
      Vegetables & 6,155 & 7,235 & 45 & 58 \\
      Condiments & 3,737 & 4,084 & 27 & 30 \\ 
      Breads & 2,515 & 2,777 & 24 & 30 \\ \hline
  \end{tabular}
  \caption{The statistics of weakly labeled dataset.}
  \label{tab:stat}
  \end{center}
\end{table}

\begin{table*}[htp]
  \small
  \begin{center}
    \setlength{\tabcolsep}{1.7mm}{
    \begin{tabular}{|c|ccc|ccc|ccc|ccc|ccc|} \hline 
        & \multicolumn{3}{c|}{\textbf{CY}} & \multicolumn{3}{c|}{\textbf{BN}} & \multicolumn{3}{c|}{\textbf{YO}} & \multicolumn{3}{c|}{\textbf{MN}} & \multicolumn{3}{c|}{\textbf{ARZ}} \\
        & \textbf{P} & \textbf{R} & \textbf{F1} & \textbf{P} & \textbf{R} & \textbf{F1} & \textbf{P} & \textbf{R} & \textbf{F1} & \textbf{P} & \textbf{R} & \textbf{F1} & \textbf{P} & \textbf{R} & \textbf{F1} \\ \hline
      CNN-CRFs & 84.4 & 76.2 & 80.1 & 92.0 & 89.1 & 90.5 & \textbf{80.9} & 68.9 & 74.4 & 87.3 & 85.5 & 86.3 & 88.6 & 86.7 & 87.6\\
      BiLSTM-CRFs & \textbf{86.0} & 77.8 & 81.6 & 93.3 & 91.5 & 92.3 & 74.1 & 68.9 & 71.3 & \textbf{89.0} & 85.5 & 87.1 & \textbf{89.5} & 88.5 & 89.0 \\
      Trans-CRFs & 83.7 & 73.2 & 78.1 & 93.0 & 85.9 & 89.3 & 80.2 & 60.5 & 69.0 & 88.0 & 80.0 & 83.8 & 88.9 & 83.2 & 85.9 \\ \hline
      BiLSTM-PCRFs & 85.2 & 79.6 & 82.3 & 91.2 & 92.7 & 91.9 & 68.1 & 70.2 & 69.1 & 82.5 & 91.2 & 86.6 & 84.0 & 90.7 & 87.1 \\ \hline
      Ours & 82.8 & \textbf{82.5} & \textbf{82.6} & \textbf{93.4} & \textbf{93.5} & \textbf{93.4} & 73.5 & \textbf{76.8} & \textbf{75.1} & 86.9 & \textbf{93.6} & \textbf{90.1} & 87.7 & \textbf{91.5} & \textbf{89.5} \\ \hline
      
  \end{tabular}}
  \caption{Performance (\%) on low-resource languages.}
  \label{tab:lang}
  \end{center}
\end{table*}

We use 20190120 Wikipedia dump for WL data construction, where the ratio of words in anchors to the whole sentence is nearly 0.12, 0.07, 0.14, 0.07 and 0.06 for languages CY, BN, YO, MN and ARZ, and 0.13 for food domain, demonstrating that unlabeled words are dominant. By heuristically setting $\theta_q=0.1,\theta_n=0.9$, we obtain 56,571, 16,718, 4,131, 8,332, 6,266, 11,297 high-quality and 49,970, 50,197, 32,417, 10,918, 12,434, 16,501 noisy WL sentences for language CY, BN, YO, MN and ARZ, and food domain, respectively. For correctness, we then pick up test data of 25\% sentences that has highest annotation confidence and exceed 0.3 coverage. We randomly choose 25\% of high-quality data as validation for early stop, and the rest for training. The statistics\footnote{The statistics includes noisy data, which greatly increases the size but cannot be used for evaluation.} is in Table~\ref{tab:stat}.

\vpara{Training Details}

For tuning of hyper-parameters, we set non-entity feature weights to $\lambda_1=0, \lambda_2=0.9, \lambda_3=0.1$ heuristically. We pre-train word embeddings using Glove~\cite{pennington2014glove}, and fine-tune embeddings during training. We set the dimension of words and characters as 100 and 30, respectively. We use 30 filter kernels, where each kernel has the size of 3 in character CNN, and dropout rate is set to 0.5. For bi-LSTM, the hidden state has 150 dimensions. The batch size is set to 32 and 64 for sequence labeling module and classification module. We adopt Adam with L2 regularization for optimization, and set the learning rate and weight decay to $0.001$ and $1e^{-9}$.

\vpara{Baselines}
Since most low-resource name tagging methods introduce external knowledge (Section~\ref{sec:rw}), which has limited availability and is out of the scope for this paper, we arrive at two types of baselines from weakly supervised models:

\textbf{Typical NN-CRFs models}~\cite{ni2017weakly} by selecting high-quality WL data and regarding unlabeled words as O, which usually achieve very competitive results. NN denotes CNN, Transformer (Trans for short) or Bi-LSTM.

\textbf{NN-PCRFs model}~\cite{yang2018distantly,shang2018learning}. Although they achieves state-of-the-art performance, methods of this type are only evaluated in specific domains and require a small set of seed annotations or a domain dictionary. We thus carefully adapt them to low-resource languages and domains by selecting the highest-quality WL data ($\theta_n>0.3$) as seeds\footnote{We adopt the common part of their models related to handling weakly labeled data, removing the other parts that are specifically designed for domains, such as instance selector~\cite{yang2018distantly} which makes it worse since we have already selected the high-quality data.}.

\subsection{Results on Low-Resource Languages}
\label{sec:lrl}
Table~\ref{tab:lang} shows the overall performance of our proposed model as well as the baseline methods (P and R denote Precision and Recall). We can see:

Our method consistently outperforms all baselines in five languages w.r.t F1, mainly because we greatly improve recall (2.7\% to 9.34\% on average) by taking best advantage of WL data and being robust to noise via two modules. As for the precision, partial-CRFs perform poorly compared with CRFs due to the uncertainty of unlabeled words, while our method alleviates this issue by introducing linguistic features in non-entity sampling. An exception occurs in CY, because it has the most training data, which may bring more accurate information than sampling. Actually, we can tune hyper-parameter non-entity ratio $\alpha$ to improve precision\footnote{In this table, we show the performance using the same hyper-parameters in different languages for fairness.}, more studies can be found in Section~\ref{sec:nsr}. Besides, the sampling technique can utilize more prior features if available, we leave it in future.

Among all encoders, Bi-LSTM has greater ability for feature abstraction and achieves the highest precision in most languages. An unexpected exception is Yoruba, where CNN achieves the higher performance. This indicates that the three encoders capture textual semantics from different perspectives, thus it is better to choose the encoder by considering the linguistic natures.

\begin{figure*}[htb]
  \centering
  \subfigure[Efficiency analysis.]{
  \label{fig:EA}\includegraphics[width=0.34\textwidth]{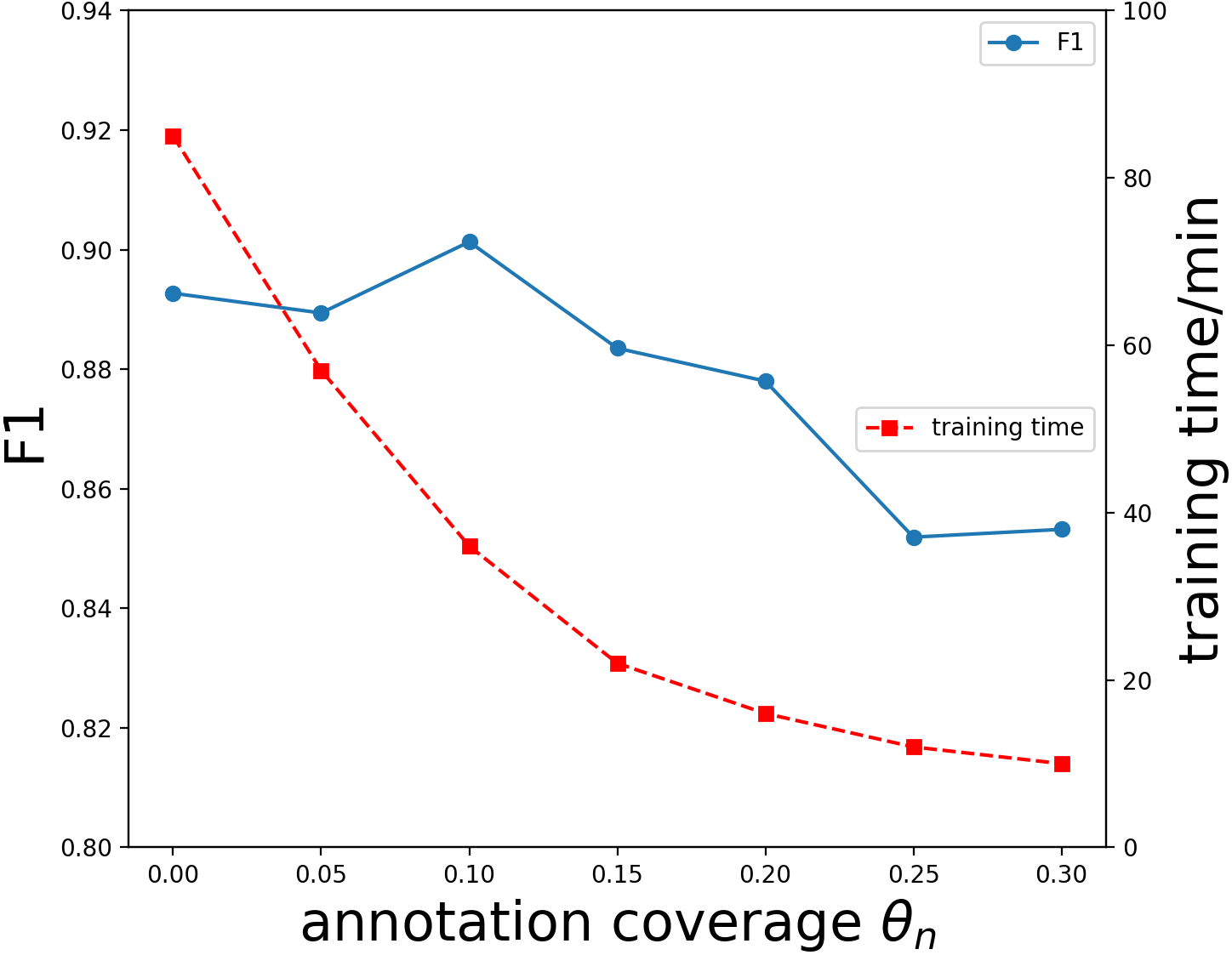}}
  \subfigure[Impact of non-entity sampling ratio.]{
  \label{fig:ne_ratio}\includegraphics[width=0.32\textwidth]{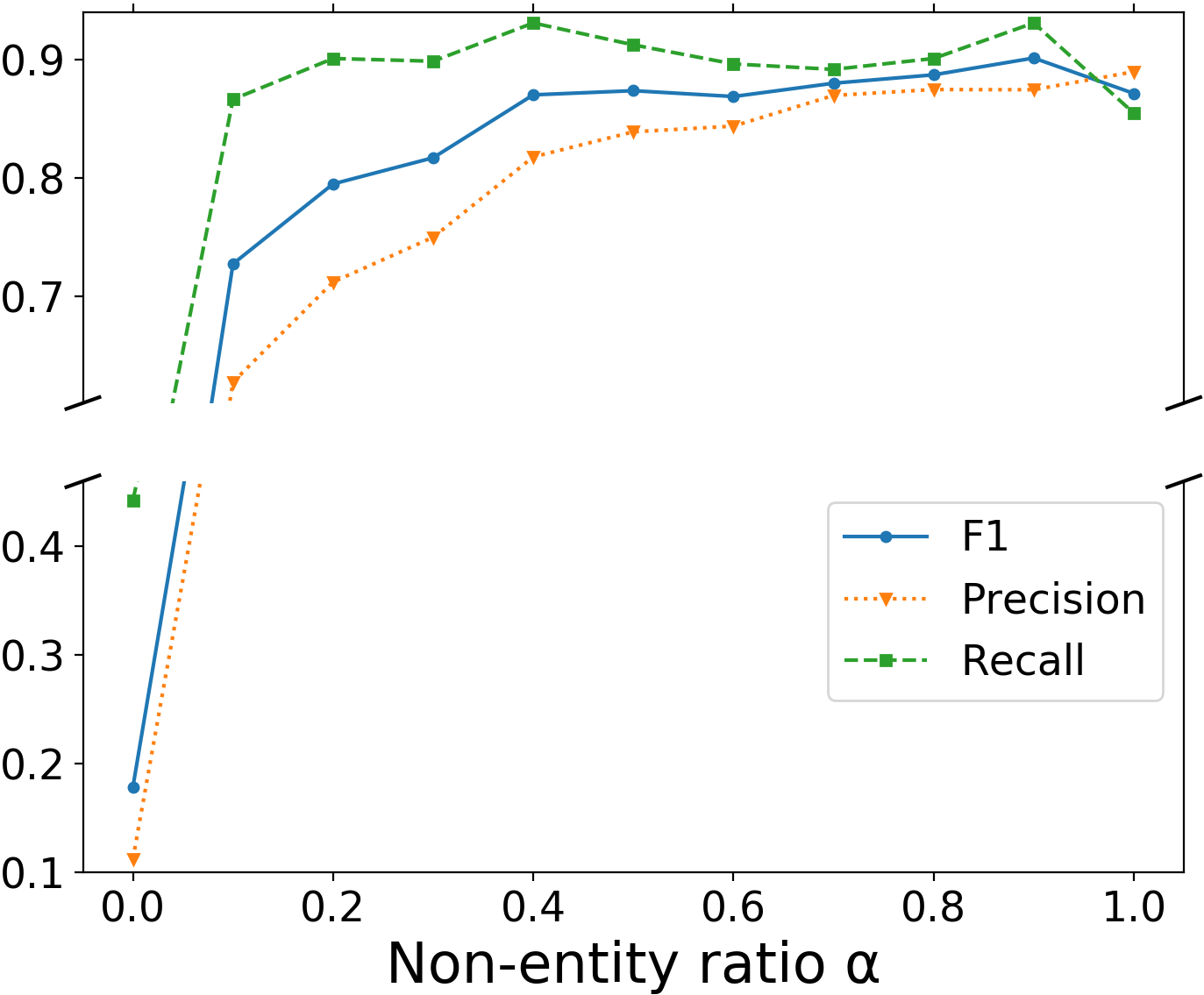}}
  \subfigure[Impact of non-entity features.]{
  \label{fig:ne_fea}\includegraphics[width=0.31\textwidth]{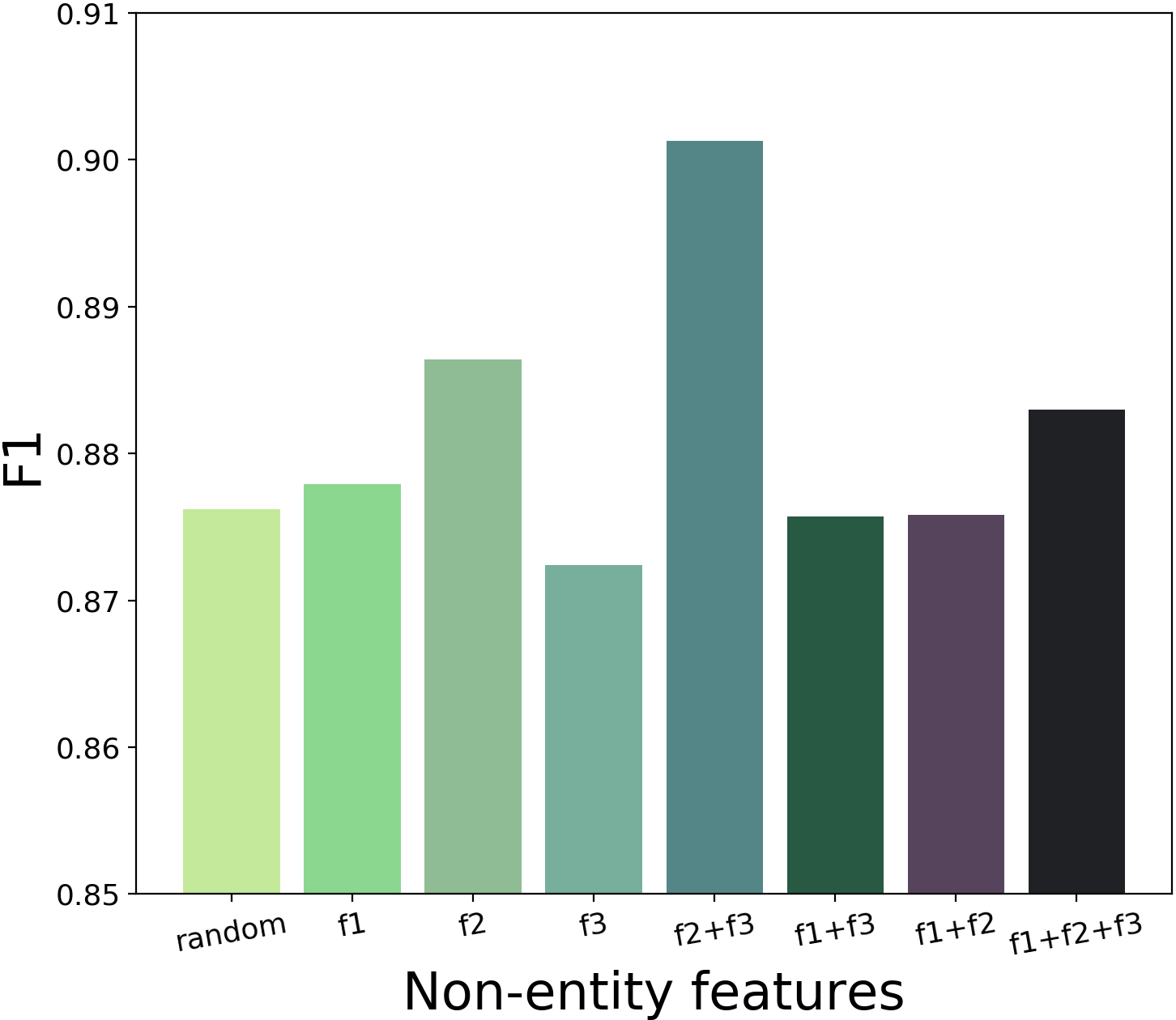}}
  \caption{Ablation study of our model in Mongolian.}
  \label{fig:ablation}
\end{figure*}

As for the impacts of resources, all the models perform the worst in Yoruba. Interestingly, we conclude that the performance for name tagging in low-resource languages doesn't depend entirely on the absolute number of mentions in the training data, but largely on the average number of annotations per sentence. For example, Bengali has $1.9$ mentions per sentence and all methods achieve their best results, while the opposite is Welsh with $1.4$ mentions per sentence. This verifies our data selection scheme (e.g., annotation coverage $n(\cdot)$), and we will give more discussion in Section~\ref{sec:ea}.

\subsection{Results on Low-Resource Domains}

\begin{table}[htp]
  \small
  \begin{center}
    \setlength{\tabcolsep}{1.5mm}{
    \begin{tabular}{|c|ccccc|c|} \hline 
        & \multicolumn{6}{c|}{\textbf{Food}} \\ \cline{2-7}
        & \textbf{D} & \textbf{M} & \textbf{V} & \textbf{C} & \textbf{B} & \textbf{All}\\ \hline
      CNN-CRFs & 67.8 & 69.8 & 57.9 & 42.8 & 46.5 & 60.9 \\
      BiLSTM-CRFs & 64.9 & 69.0 & 62.8 & 50.0 & 62.2 & 63.5 \\
      Trans-CRFs & 62.1 & 68.9 & 59.6 & 43.4 & 54.5 & 60.6 \\ \hline
      BiLSTM-PCRFs & 66.1 & 70.7 & 67.2 & 44.4 & 58.3 & 64.4 \\ \hline

      Ours & \textbf{72.0} & \textbf{76.4} & \textbf{70.0} & \textbf{53.5} & \textbf{66.6} & \textbf{70.1} \\ \hline
      
  \end{tabular}}
  \caption{F1-score (\%) on food domain.}
  \label{tab:food}
  \end{center}
\end{table} 

Table~\ref{tab:food} shows the overall performance in food domain, where D, M, V, C and B denote: Drink, Meat, Vegetables, Condiments and Breads.
We can observe that there is a performance drop compared to that in low-resource languages, mainly because of more types and sparse training data. Our model outperforms all of the baselines in all food types by 7.8\% on average. The performance in condiments is relatively low, because most of them are composed of meat or vegetables, such as steak sauce, which is overlapped with other types and make the recognition more difficult.

\begin{figure}[htb]
  \centerline{\includegraphics[width=0.48\textwidth]{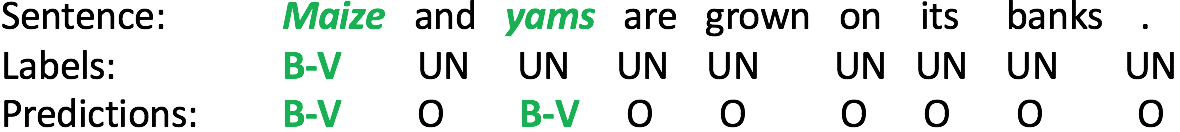}}
  \caption{Our predictions on a noisy WL sentence.}
  \label{fig:case}
\end{figure}

Here is a representative case demonstrating that our model is robust to noise induced by unlabeled words.
In Figure~\ref{fig:case}, the sentence is from the noisy WL training data of food domain, and only \textit{Maize} is labeled as B-V. Although our model is trained on this sentence, it successfully predicts \textit{yams} as B-V. This example shows that our two-modules design can utilize the noisy data while avoiding side effects caused by incomplete annotation.

\subsection{Efficiency Analysis}
\label{sec:ea}
We utilize $\theta_n$, the main factor to annotation quality (Section~\ref{sec:lrl}), to trade off between high-quality and noisy WL data. As shown in Figure~\ref{fig:EA}, the red curve denotes the training time and the blue curve denotes F1. We can see that the performance of our model is relatively stable when $\theta_n\in[0,0.15)$, while the time cost drops dramatically (from 90 to 20 minutes), demonstrating the robustness and efficiency of two-modules design. When $\theta_n\in[0.15,0.3]$, the performance decreases greatly due to less available high-quality data for sequence labeling module; meanwhile, little time is saved through classification module. Thus, we pick up $\theta_n=0.1$ in experiments. A special case happens when $\theta_n=0$, our model degrades to sequence labeling without pre-trained classifier. We can see the performance is worse than that of $\theta_n=0.1$ due to massive noisy data.

\subsection{Impact of Non-Entity Sampling Ratio}
\label{sec:nsr}
We use non-entity ratio $\alpha$ to control sampling, and a higher $\alpha$ denotes that more unlabeled words are labeled with O. As shown in Figure~\ref{fig:ne_ratio}, the precision increases as more words are assigned with labels, while the recall achieves two peaks ($\alpha=0.4,0.9$), leading to the highest F1 when $\alpha=0.9$, which conforms to the statistics in~\citet{augenstein2017generalisation}. There are two special cases. When $\alpha=0$, our model degrades to a NN-PCRFs model without non-entity sampling and there is no seed annotations for training. We can see the model performs poorly due to the dominant unlabeled words (Section~\ref{sec:nes}). When $\alpha=1$ indicating all unlabeled words are sampled as O, our model degrades to NN-CRFs model, which has higher precision at the cost of recall. Clearly, the model suffers from the bias to O labeling.

\subsection{Impact of Non-Entity Features}
We propose three features for non-entity samples: nearby entities ($f_1$), ever within entities ($f_2$) and term/document frequency ($f_3$). We now investigate how effective each feature is. Figure~\ref{fig:ne_fea} shows the performance of our model that samples non-entity words using each feature as well as their combinations. The first bar denotes the performance of sampling without any features. It is not satisfying but competitive, indicating the importance of non-entity sampling to partial-CRFs. The single $f_2$ contributes the most, and gets enhanced with $f_3$ because they provide complementary information. Surprisingly, $f_1$ seems better than $f_3$, but makes the model worse if we use it combined with $f_2,f_3$, thus we set $\lambda_1=0$.

\section{Conclusions}

In this paper, we propose a novel name tagging model that consists of two modules of sequence labeling and classification, which are combined via shared parameters. We automatically construct WL data from Wikipedia anchors and split them into high-quality and noisy portions for training each module. The sequence labeling module focuses on high-quality data and is costly due to the partial-CRFs layer with non-entity sampling, which models all possible label combinations. The classification module focuses on the annotated words in noisy data to pretrain the tag classifier efficiently. The experimental results in five low-resource languages and a specific domain demonstrate the effectiveness and efficiency.

In the future, we are interested in incorporating entity structural knowledge to enhance text representation~\cite{cao2017bridge,cao2018joint}, or transfer learning~\cite{sun2019meta} to deal with massive rare words and entities for low-resource name tagging, or introduce external knowledge for further improvement.

\section*{Acknowledgments}

NExT++ research is supported by the National Research Foundation, Prime Minister's Office, Singapore under its IRC@SG Funding Initiative.

\bibliography{ms}
\bibliographystyle{acl_natbib}

\end{document}